\documentclass{article}
\usepackage{ijcai22}

\usepackage{times}
\usepackage{soul}
\usepackage{url}
\usepackage[hidelinks]{hyperref}
\usepackage[utf8]{inputenc}
\usepackage[small]{caption}
\usepackage{graphicx}
\usepackage{amsmath}
\usepackage{amsthm}
\usepackage{booktabs}
\usepackage{algorithm}
\usepackage{algorithmic}

\usepackage{epsfig}
\usepackage{graphicx}
\usepackage{amsmath}
\usepackage{amssymb}
\usepackage{booktabs}
\usepackage{comment}

\usepackage{multirow}

\usepackage{booktabs}
\usepackage{array, caption, threeparttable}
\usepackage{caption}

\usepackage{algorithm}
\usepackage{listings}
\usepackage{color}

\usepackage{mathtools}

\usepackage{todonotes}
\usepackage{gensymb}

\usepackage{subfigure}

\title{Rethinking Feature Uncertainty in Stochastic Neural Networks for\\ Adversarial Robustness}

\author{
Hao Yang$^1$
\and
Min Wang$^1$\and
Zhengfei Yu$^1$\and
Yun Zhou$^1$\thanks{Corresponding author. This work is supported by Huxiang Youth Talent Support Program (No. 2021RC3076) and Training Program for Excellent Young Innovators of Changsha (No. KQ2009009).}
\\
\affiliations
$^1$National university of Defense Technology
\emails
\{yanghao,wangminwm,yuzhengfei19, zhouyun\}@nudt.edu.cn}

\begin{document}
\maketitle
\begin{abstract}
It is well-known that deep neural networks (DNNs) have shown remarkable success in many fields. However, when adding an imperceptible magnitude perturbation on the model input, the model performance might get rapid decrease. To address this issue, a randomness technique has been proposed recently, named Stochastic Neural Networks (SNNs). Specifically, SNNs inject randomness into the model to defend against unseen attacks and improve the adversarial robustness. However, existed studies on SNNs mainly focus on injecting fixed or learnable noises to model weights/activations. In this paper, we find that the existed SNNs performances are largely bottlenecked by the feature representation ability. Surprisingly, simply maximizing the variance per dimension of the feature distribution leads to a considerable boost beyond all previous methods, which we named maximize feature distribution variance stochastic neural network (MFDV-SNN). Extensive experiments on well-known white- and black-box attacks show that MFDV-SNN achieves a significant improvement over existing methods, which indicates that it is a simple but effective method to improve model robustness. 
\end{abstract}

\section{Introduction}
Deep learning models have shown promising results in many fields. However, a small perturbation on the input, which is hard to identify by human eye, can subvert the model's predictions easily. The perturbed data is called the adversarial example~\cite{goodfellow2014explaining}. This phenomenon that attack leads to a decrease in model performance has attracted many researchers and security service organizations to build safer and more secure models. 

Among the recently developed defense methods, Stochastic Neural Networks (SNNs) have shown great potentials for building robust models. As we all know, stochastic noise helps models avoid getting trapped into local minima during training phases~\cite{hopfield1982neural,srivastava2014dropout}. Within the aforementioned characteristics of stochastic noise, SNNs transform deterministic models into stochastic models by injecting randomness noises into model activations/weights. Within more uncertainty, the model can explore more profound network representation. When the model converges, the uncertainty of the noise will force the model to be in a stable state with relatively locally optimal. Thus, it enhances robustness of the model. 

Inspired by the above analysis, a natural idea was came up: \textbf{Does larger uncertainty lead to higher robustness?} In this paper, we construct the SNN model via injecting isotropic noise into the feature layer, and further explore the feature uncertainty for adversarial robustness in SNNs. It is exciting that simply maximizing per dimension of the feature distribution brings huge model robustness. It is also worth emphasizing that currently most defense methods are based on adversarial training - a data argumentation method that uses adversarial examples generated by the attack method to retrain the networks~\cite{kurakin2016adversarial,madry2017towards}. Although adversarial training is simple and proved effective for improving model robustness, it significantly leads to higher computational costs and more training time. Moreover, mixed clean examples and adversarial examples might ``obfuscate'' the clean data gradient information. Thus, improving adversarial robustness is at the expense of clean data accuracy. 

On the contrary, our model does not require adversarial training to achieve robustness, which means it does not require high computation costs and degrade the clean data accuracy. To sum up, our contributions are as follows:
\begin{itemize}
	\item We propose a simple and efficient stochastic neural network which maximize the feature distribution variance (MFDV-SNN)  for improving adversarial robustness.
	\item To the best of our knowledge, we are the first to explore the effect of feature uncertainty in SNNs on adversarial robustness.
	\item The proposed method does not require adversarial training and can maintain the strong representational capability while requiring only a low computational cost.
	\item Extensive experiments on white- and black-box attacks show our proposed method is compelling.
\end{itemize}

\section{Related Work}
\subsection{Adversarial Attack}
Many methods have been proposed to attack deep learning models. Researchers further divided attack methods into white- and black-box attacks according to whether they can obtain the gradient information.

\noindent\textbf{White-box attack:}\quad
 White-box attacks mean that the attacker knows the model gradient information. A simple yet effective white-box attack method is called Fast Gradient Sign Method (FGSM)~\cite{goodfellow2014explaining}, which adds a small perturbation in the direction of the sign of the gradient updates. It can be formulated as 
\begin{equation}
\vec{x}^{\prime}=\vec{x}+\epsilon \cdot \operatorname{sign}\left(\nabla_{\vec{x}} \mathcal{L}(h(\vec{x}), y)\right)
\label{eq1}
\end{equation}
where $\vec{x}$ denotes the input image, $\epsilon$ denotes the perturbation strength, $\mathcal{L}$ denotes the loss function and $h(\cdot)$ denotes the target model. Kurakin further update the one-step attack FGSM to multi-step attack which named Basic Iterative Method (BIM)~\cite{kurakin2016adversarial}. Compared to FGSM, BIM uses a smaller step size to explore the possible adversarial direction. Furtherly, Madry updates BIM by random initializing the input point, which is one of the most strongest first-order attack named Projected Gradient Descent (PGD)~\cite{madry2017towards}. It can be formulated as 
\begin{equation}
\vec{x}^{t+1}=\Pi_{\vec{x}+s}\left(\vec{x}^{t}+\alpha \cdot {sgn}\left(\nabla_{\vec{x}} \mathcal{L}\left(h(\vec{x}^{t}, y\right)\right)\right.
\end{equation}
where $\Pi_{\vec{x}+s}$ is the projection operation which force the adversarial example in the $\ell_{p}$ ball $s$ around $\vec{x}$, and $\alpha$ is the step size. Another strong first-order attack algorithm is called C$\&$W attack~\cite{carlini2017towards} which finds adversarial example by solving the following optimization function formulated as 
\begin{equation}
\min \left[\|\delta\|_{p}+c \cdot h(\vec{x}+\delta)\right] \text { s.t. } \vec{x}+\delta \in[0,1]^{n}
\end{equation}
where $p$ is the norm distance, commonly choosing from $\{0,2, \infty\}$.

\noindent\textbf{Black-box attack:}\quad 
Unlike white-box attacks, black-box attackers can only access the model through queries. There are mainly two ways to fool a model. One is to train a substitute of the model~\cite{papernot2017practical}, in which attackers query from the target model and generate a synthesized dataset with input and the corresponding output. Due to the transferability of adversarial examples, attackers can attack alternative models and target models.The limitation of this method is that it cannot execute multiple queries in reality. The other is to estimate the gradients via multiple queries to the targeted model~\cite{su2019one}. Among them, zero-order optimization~\cite{chen2017zoo} algorithms aim to estimate the gradients of the target model directly. 

\subsection{Stochastic Defense}
Nowadays, SNNs have shown promising results via injecting fixed or learnable noise into model activations/weights. RSE~\cite{liu2018towards} uses ensemble tricks to improve model robustness. Specifically, they inject Gaussian noise to multi-layers during training and then perform multiple forward passes to test it. It means they only need training one time and can be viewed as an ensemble model. RSE mainly uses fixed variance by hand-tuned. Parameter noise injection (PNI)~\cite{he2019parametric} further proposed to learn a sensitive parameter to control the variance. Moreover, L2P~\cite{jeddi2020learn2perturb} updates PNI by alternating training a noise module and a neural network module, which they called "alternating back-propagation." 
\subsection{Connection to Similar Work}
In this section, we mainly discuss some similar but different works to clarify the contribution of our work. The most related work is proposed by Yu~\cite{yu2019simple}, in which they propose to use a max-margin entropy loss to regularize the feature distribution. However, we need to highlight some main differences between them. Firstly, the construct way of the Gaussian layer is different. In their work, a parallel Gaussian mean and Gaussian variance may meet mistakes when increasing the Gaussian variance. In this situation, the Gaussian mean and Gaussian variance share the same parameters on the feature layer but different optimization ambitions which may pass on obfuscate gradient information to the last feature layer matrix through back-propagation. Secondly, the margin $b$ in their paper largely restricted the model representation ability, and it means the authors given a solid prior information which can not ensure rationality. A similar situation can be seen in the deep variational information bottleneck (VIB)~\cite{alemi2016deep} model that uses a definite Gaussian variance, which also restricts the representation ability for model robustness.

To sum up, while previous researches often use fixed feature uncertainty, we highlight that we are the first to explore the relationship between feature uncertainty and model robustness. The key finding of our research is that larger uncertainty will bring higher robustness. That is why the proposed method MFDV-SNN achieves state-of-the-art results compared to previous various SNN-based models. Moreover, the proposed method is simple and efficient, motivating researchers to rethink the feature uncertainty in SNNs for adversarial robustness.

In practice, we only need to maximize the unbounded feature distribution variance to achieve significant improvement. The unbounded high variance will be self-adaptive to the model architecture and dataset and then explore a more stable representation. We also need to claim that the unbounded variance will not collapse since the gradient is $-\frac{1}{\sigma}$, the over-large variance will not back-propagation gradient information to update the network parameter. Extensive experiments on white- and black-box attacks confirm that the proposed key point is important enough.

\section{Methodology}
\begin{figure}
	\centering
	\includegraphics[width=\linewidth]{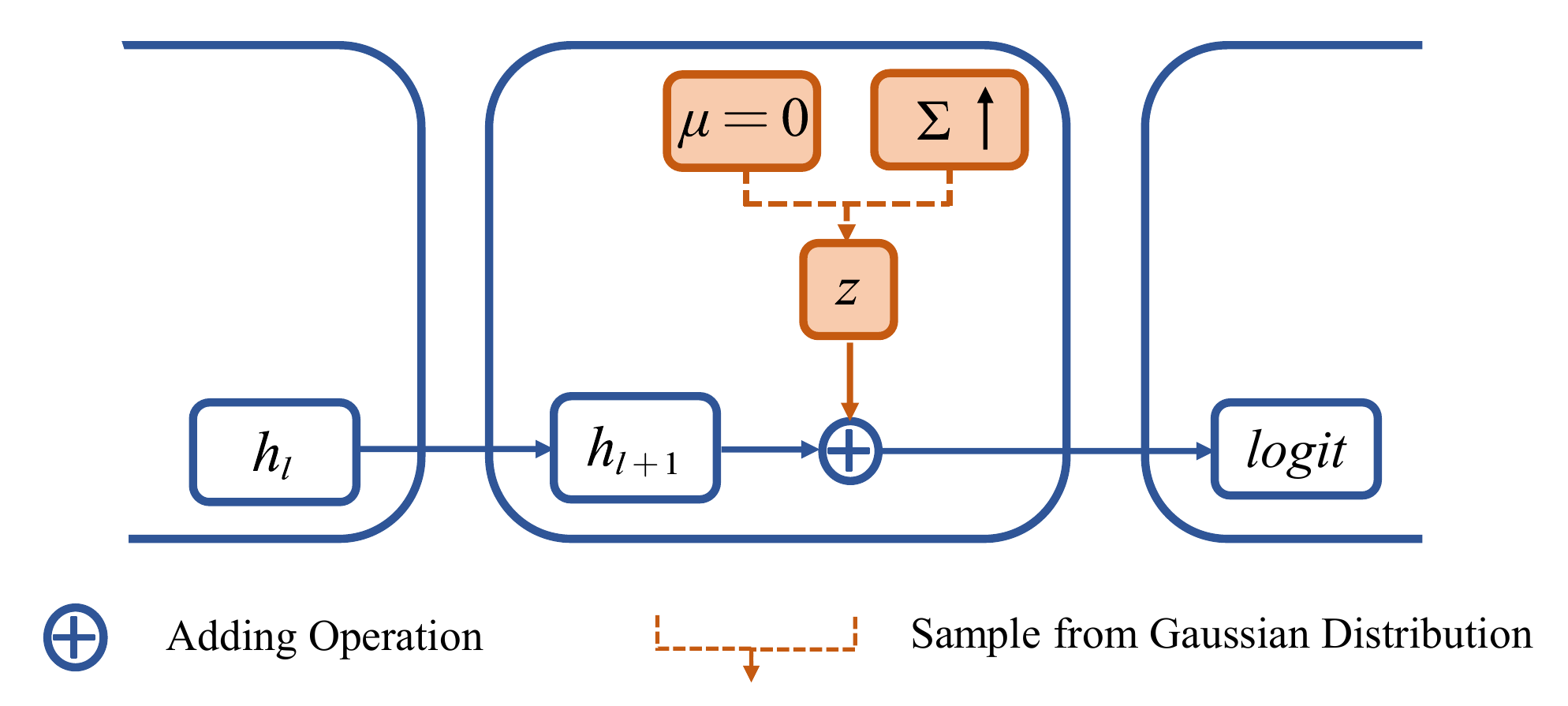}
	\caption{An illustration of our proposed MFDV-SNN.}
\label{fig:flow}
\end{figure}
In this section, we introduce in detail the implementation process of the proposed MFDV-SNN. The critical point is shown in Figure~\ref{fig:flow}.
\subsection{Stochastic layer}
As illustrated in Figure~\ref{fig:flow}, we take the last three-layer as a detailed description. The data passes through a neural network to get $h_{l}$, then $h_{l}$ pass through a layer which is usually a linear layer and gets the feature extractor $h_{l+1}$ . We do not directly build a parallel Gaussian mean and Gaussian variance layer as~\cite{yu2019simple} since increasing the variance may obfuscate the original feature layer.
Instead, we keep the original feature plus a maximum variance Gaussian distribution, where the Gaussian mean is a zero matrix. Thus, the $h_{l+1}$ can not only maintain the original feature representation ability but also obtain uncertainty via injected noise. The idea behind this is that the original feature can preserve the data manifold well for downstream tasks such as image classification. However, it is not appropriate within the adversarial setting, as a large proportion of information is redundant and even harmful to model robustness. Thus, we need considerable uncertainty to explore a more robustness representation. What is more, we use a non-information prior to initializing the Gaussian variance. In practice, we sample the same dimension with the feature dimension from the uniform distribution. Then, we establish a Gaussian distribution $z$, and sample from the Gaussian distribution plus with the original hidden representation $h_{l+1}$. Finally, get the logit of the network.

\subsection{Loss Function}
As discussed above, we need to maximize the feature uncertainty during model training. The simplest implementation is to add the built Gaussian variance to the loss function. It needs to emphasize that we do not directly assign a parametric variance, e.g., var= 20. One reason is that this will hinder the model convergence, and it may lead to model to collapse at the training beginning. Another reason is that we do not know the deterministic uncertainty that model needs, which is relevant to model architecture and the datasets. 

Instead, we initialize the Gaussian variance with the uniform distribution values from zero to one. As the network is trained, the variance increases gradually at small steps $\Delta$. The optimization process will iterate dynamically, which means that once the model converges to local minima, greater uncertainty will force it to explore a more stable state. Finally, it will adapt to a more robust state automatically.

Thus, the loss function can be formulated as
\begin{equation}
\mathcal{L}=\mathcal{L}_{C}-\lambda_{1} \sum_{i=1}^{D} \ln \left(\vec{\sigma}_{i}   +\Delta_{i} \right)+ \lambda_{2} \vec{w}^{T} \vec{w}
\label{losses}
\end{equation}
where $\mathcal{L}_{C}$ is the cross-entropy loss, and D denotes the feature dimension of the penultimate layer. While adding all the dimensions of the variance simply may meet our requirements,  we take the $\ln\left(\vec{\sigma}_{i}\right)$ operation. It has two advantages. One is to facilitate the derivative calculation of the gradient. The other is to slow down the numerical change of variance. In particular, although we emphasize that the network will not collapse, the sudden and significant change in the variance value at the beginning of the network training is not conducive to the back propagation of the gradient. The $\ln \left(\vec{\sigma}_{i}\right)$ operation can be used as an active protection measure. The last item is $L_{2}$ regularization to penalty weights over-large, $\lambda_{1}$ controls the power of variance, $\lambda_{2}$ controls the power of weights penalty.

\section{Experiments}
\subsection{Datasets \& Adversarial Attacks}
Three well-known datasets are used in our experiments: SVHN, CIFAR-10, and CIFAR-100. The SVHN dataset consists of 73K training data and 26K testing data of size 32x32x3 with ten classes. The CIFAR-10 dataset consists of 50K training data and 10K testing data of size 32x32x3 with ten classes. For the CIFAR-100 dataset, the size of training data and testing data is the same as CIFAR-10, but with one-hundred classes. The attack algorithms in our experiments contain white- and black-box attacks. For white-box attacks, we use FGSM, PGD$_{10}$, C\&W, and stronger PGD$_{100}$ to evaluate the efficiency of the proposed MFDV-SNN. For Black-box attacks, n-Pixel attacks and more powerful Square attack are used in our experiment.

\subsection{Experimental Setup}
The main backbone in our experiment is ResNet-18~\cite{he2016deep}, the attacks follows~\cite{he2019parametric,jeddi2020learn2perturb,eustratiadis2021weight}. In Table~\ref{attack1} and Table~\ref{attack2}, the attack strength $\epsilon$ of FGSM and PGD is $8/255$. For PGD attack, the steps we set $k=10$ and the $\alpha$ we set to $\epsilon/10$, which follows~\cite{he2019parametric,eustratiadis2021weight}. For C\&W attack, the learning rate we set $\alpha=5 \cdot 10^{-4}$, the number of iterations $k=1000$, initial constant $c=10^{-3}$ and maximum binary steps $b_{max}=9$ same as \cite{jeddi2020learn2perturb,eustratiadis2021weight}. For the n-Pixel attack, we set the population size $N=400$ and maximum number $k_{max}=75$ same as \cite{jeddi2020learn2perturb}, and we conduct a stronger 5-pixel attack which is not implemented in their setting. For Square attack, we refer to the implementation from~\cite{croce2020reliable}. All experiments are performed on the Pytorch platform of version 1.7.0.

\subsection{Comparison to Prior Stochastic Defenses}
In this section, we list the comparative stochastic defense methods. \textbf{Adv-BNN}~\cite{liu2018adv}: A combination of Bayesian neural network with adversarial training. \textbf{PNI}~\cite{he2019parametric}: Injecting Gaussian noise to multilayers. \textbf{L2P}~\cite{jeddi2020learn2perturb}: Updating PNI by learning a perturbation injection module and alternating training the noise and network module. \textbf{SE-SNN}~\cite{yu2019simple}: Introducing a margin entropy loss. What is more, there are partial comparisons against \textbf{WCA-Net}~\cite{eustratiadis2021weight} and \textbf{IAAT}~\cite{xie2019feature}.

\begin{table}[t]
\centering
\scalebox{0.9}
{\begin{tabular}{cccc}
\toprule
Method & Clean &FGSM &PGD\\
\midrule
Adv-BNN~\cite{liu2018adv} & 82.2 &60.0 &53.6\\
PNI~\cite{he2019parametric} & 87.2 & 58.1 & 49.4\\
L2P~\cite{jeddi2020learn2perturb} & 85.3 &62.4 & 56.1\\
SE-SNN~\cite{yu2019simple} & 92.3& 74.3 &-\\
WCA-Net~\cite{eustratiadis2021weight} & 93.2 &77.6 &71.4\\
\textbf{MFDV-SNN (Ours)} &\textbf{93.7}&\textbf{85.7}&\textbf{79.6}\\
\bottomrule
\end{tabular}}
\caption{Comparison of state-of-the-art SNNs for FGSM and PGD attacks on CIFAR-10 with a ResNet-18 backbone.}
\label{attack1}
\end{table}

\begin{table}[t]
\centering
\scalebox{1.0}
{
\begin{tabular}{cccc}
\toprule
Method & Clean &FGSM &PGD\\
\midrule
Adv-BNN~\cite{liu2018adv} & 58.0 &30.0 &27.0\\
PNI~\cite{he2019parametric} & 61.0 & 27.0 & 22.0\\
L2P~\cite{jeddi2020learn2perturb} & 50.0 &30.0 & 26.0\\
IAAT~\cite{xie2019feature}& 63.9 &- & 18.5\\
\textbf{MFDV-SNN (Ours)} &\textbf{69.4}&\textbf{47.1}&\textbf{37.3}\\
\bottomrule
\end{tabular}
}
\caption{Comparison of state-of-the-art SNNs for FGSM and PGD attacks on CIFAR-100 with a ResNet-18 backbone.}
\label{attack2}
\end{table}

\begin{table}[t]
\scalebox{0.98}
{
\begin{tabular}{@{}cccccc@{}}
\toprule
&Strength & Adv-BNN & PNI  & L2P  & \textbf{MFDV-SNN} \\
\midrule
& Clean   & 82.2    & 87.2 & 85.3   &\textbf{93.7}      \\
\midrule
\multirow{4}{*}{\rotatebox[origin=c]{0}{C\&W}}
& k=0   & 78.9    & 66.9 & 83.6     & \textbf{88.8}     \\
& k=0.1  & 78.1    & 66.1 & 84      & \textbf{87.9}      \\
& k=1   & 65.1    & 34   & 76.4 & \textbf{87.2} \\
& k=2   & 49.1    & 16   & 66.5    & \textbf{86.6}      \\
& k=5   & 16      & 0.08 & 34.8     & \textbf{83.5}     \\
\bottomrule
\end{tabular}
}
\caption{Comparison of state-of-the-art SNNs for white-box C$\&$W attack on CIFAR-10 with a ResNet-18 backbone. }
\label{cw}
\end{table}

\begin{table}[htbp]
\centering
\begin{tabular}{ccccccc}
\toprule
&Strength & Adv-BNN &PNI &L2P&\textbf{MFDV-SNN}\\
\midrule
\multirow{4}{*}{\rotatebox[origin=c]{90}{n-Pixel}}
&Clean & 82.2 &87.2 &85.3&\textbf{93.7}\\
&1 pixel &68.6 & 50.9 & 64.5&\textbf{85.4}\\
&2 pixels & 64.6 &39.0 & 60.1&\textbf{80.4}\\
&3 pixels & 59.7 &35.4 &53.9 &\textbf{76.0}\\
&5 pixels &- &- &- &\textbf{68.0}\\
\bottomrule
\end{tabular}
\caption{Comparison of state-of-the-art methods for black box n-Pixel attack on CIFAR-10 with a ResNet-18 backbone.}
\label{pixel_attack}
\end{table}

\noindent\textbf{Against White-box Attacks.}\quad
In this section, we mainly focus on evaluating the proposed MFDV-SNN against white-box attacks. In Table~\ref{attack1}, we compare the state-of-the-art SNNs for FGSM and PGD attacks on CIFAR-10 with a ResNet-18 backbone. The results show that the proposed MFDV-SNN outperforms all comparison algorithms. It is worth noticing that most relevant SE-SNN have not released the official code and the result of ResNet-18.  Figure 2(a) in their paper~\cite{yu2019simple} shows that SE-SNN is not so strong that the accuracy decreases so fast even under the weak FGSM attack. For a fair comparison, we extract the result from~\cite{eustratiadis2021weight}. The results show that we have about 10.4$\%$ and 11.5$\%$ improvement compared with the previous best defense method WCA-Net. Compared to SE-SNN, for the FGSM attack, we have about 15.3$\%$ improvement. Compared with other recent state-of-the-art stochastic defense methods, the proposed MFDV-SNN exceeds them significantly. What is more, Adv-BNN, PNI, and L2P for contrast involve adversarial training. However, this operation improves the model's robustness at the expense of the accuracy of clean data. Instead, the proposed MFDV-SNN need not adversarial training, which means that our method need lower computation cost and lower training time. We can see from Table~\ref{attack1} that we achieve the highest accuracy on clean data, which means the more profound network optimization point explored is conducive for neural networks on the traditional classification task.

For Table~\ref{attack2}, we choose a larger dataset CIFAR-100 with one-hundred classes. The results show that the proposed MFDV-SNN still outperforms the defense methods for contrast, and the accuracy of clean data is also the highest. For the FGSM attack, compared with the best stochastic defense methods Adv-BNN and L2P, we have an improvement of about 36.3$\%$. For PGD attack, compared with the best defense method Adv-BNN, we have about 27.6 $\%$ improvement.

For Table~\ref{cw}, the attack algorithm follows foolbox\footnote{\href{https://github.com/bethgelab/foolbox}{https://github.com/bethgelab/foolbox}}, a public attack library, to evaluate the efficiency of the proposed MFDV-SNN. The results show that the proposed MFDV-SNN significantly exceeds the existed stochastic defense methods even when the confidence level $k$ is 5. Overall, although the implementation of the proposed MFDV-SNN is simple, the improvements are significant.

\noindent\textbf{Against Black-box Attacks.}\quad
In this section, we evaluate the proposed MFDV-SNN on the well-known black-box n-Pixel attack. This attack relies on evolutionary optimization and is derivative-free. The attack strength is controlled by the number of pixels it comprises. From the results in Table~\ref{pixel_attack}, we can see that the proposed MFDV-SNN outperforms other state-of-the-art methods in all attack strengths. More specifically, for ${1-,2-,3-}$ pixel attack, compared with the best defense method Adv-BNN, the proposed MFDV-SNN has improvement about 24.5$\%$, 24.5$\%$, 27.3$\%$, respectively. We add a more potent 5-pixel attack to evaluate the proposed MFDV-SNN's efficiency further.

\begin{table}[t]
\centering
\resizebox{1.0\columnwidth}{!}
{
\begin{tabular}{ccccccccccc}
\toprule
&$\epsilon/255$ & Clean &1 &2 &4 &8 &16 &32 &64 &128\\
\midrule
\multirow{2}{*}{\rotatebox[origin=c]{90}{\scriptsize PGD$_{100}$}}& No Defense &93.0 &42.7 &12.2 &3.2 &1.5 &0 &0 &0 &0\\[1pt]
& \textbf{MFDV-SNN}&\textbf{93.6} &\textbf{85.7} &\textbf{85.1}&\textbf{83.4}&\textbf{79.2}&\textbf{63.4}&\textbf{26.4}&\textbf{9.1}&\textbf{2.4}\\[1pt]
\midrule
\multirow{2}{*}{\rotatebox[origin=c]{90}{\scriptsize Square}}& No Defense &93.0 &59.9 &24.6&2.8&0.3&0&0&0&0\\[1pt]
&\textbf{MFDV-SNN} &\textbf{93.2} &\textbf{92.7} &\textbf{92.7}&\textbf{91.8}&\textbf{83.5}&\textbf{55.2}&\textbf{18.6}&\textbf{9.4}&\textbf{6.2}\\[1pt]
\bottomrule
\end{tabular}
}
\caption{Evaluating of Our proposed method with a ResNet-18 backbone on CIFAR-10, against the white-box PGD$_{100}$ and black-box Square Attack, for different values of attack strength $\epsilon$.}
\label{stronger_attack}
\end{table}

\begin{table}[t]
	\centering
    \resizebox{0.48\textwidth}{!}{
		\begin{tabular}{ccccc}
			\toprule
			Methods & Architecture & AT & Clean & PGD \\
			\midrule
			RSE~\cite{liu2018towards}& ResNext & $\times$ & 87.5& 40.0\\
			
			DP~\cite{lecuyer2019certified}& 28-10 Wide ResNet & $\times$  & 87.0& 25.0\\
		
			TRADES~\cite{zhang2019theoretically}& ResNet-18 &$\checkmark$& 84.9& 56.6\\
		
			PCL~\cite{mustafa2019adversarial}& ResNet-110 &$\checkmark$& 91.9& 46.7\\
			
			PNI~\cite{he2019parametric}& ResNet-20 (4x) &$\checkmark$& 87.7& 49.1\\
		
			Adv-BNN~\cite{liu2018adv}& VGG-16 &$\checkmark$& 77.2& 54.6\\
			
			L2P~\cite{jeddi2020learn2perturb}& ResNet-18 &$\checkmark$& 85.3& 56.3\\
			
			MART~\cite{wang2019improving}& ResNet-18&$\checkmark$ & 83.0& 55.5\\
		
			BPFC~\cite{addepalli2020towards}& ResNet-18 &$\times$ & 82.4& 41.7\\
		
			RLFLAT~\cite{song2019robust}& 32-10 Wide ResNet &$\checkmark$& 82.7& 58.7\\
		
			MI~\cite{pang2019mixup}& ResNet-50 &$\times$ & 84.2& 64.5\\
		
			SADS~\cite{vivek2020single}& 28-10 Wide ResNet &$\checkmark$& 82.0& 45.6\\
			
			WCA-Net~\cite{eustratiadis2021weight} & ResNet-18 &$\times$ & 93.2& 71.4\\
			\midrule
			\textbf{MFDV-SNN (Ours)}& ResNet-18&\textbf{$\times$}  & \textbf{93.7}& \textbf{79.6}\\
			\bottomrule
		\end{tabular}
		}
			\caption{Comparison results of the proposed method and state-of-the-art methods in providing a robust network model. All competitors evaluate their models on the untargeted PGD attack on CIFAR-10. AT: Use of adversarial training.}
			\label{sota}
\end{table}

\noindent\textbf{Stronger Attacks.}\quad
In this section, we conduct two stronger white- and black-box attacks: PGD$_{100}$ and Square attack. For White-box PGD$_{100}$ attack, the iterative steps are set as 100, which means it uses a smaller step size to explore the adversarial example. The black-box Square attack comprises the attacked data in minor localized square-shaped updates. The results are shown in Table~\ref{stronger_attack}.

\begin{table*}[t]
	\centering
\begin{tabular}{cccccccccc}
		\toprule
		{}&\multicolumn{3}{c}{SVHN}&\multicolumn{3}{c}{CIFAR-10}&\multicolumn{3}{c}{CIFAR-100}\\
	\midrule
		Model&Clean&FGSM&PGD&Clean&FGSM&PGD&Clean&FGSM&PGD\\
		\midrule
		No Defense &94.9&18.6&5.9&92.9&21.3&2.3&68.8&12.8&1.5\\
		MFDV-SNN& 94.0&\textbf{86.1}&\textbf{82.7}&\textbf{93.7}&\textbf{85.7}&\textbf{79.6}&\textbf{69.4}&\textbf{47.1}&\textbf{37.3}\\
		\bottomrule
		\end{tabular}
			\caption{Generalization study for FGSM and PGD attacks on various datasets: SVHN, CIFAR-10 and CIFAR-100. In which we use ResNet-18 as a backbone.}
			\label{ablation_dataset}
\end{table*}

\subsection{Comparison to State-of-the-Art}
We compare MFDV-SNN with several state-of-the-art defense methods proposed in recent years. Among them, some methods are SNNs, and some are not. We present the results in Table~\ref{sota}. For a fair comparison, some results are extracted from the original paper and~\cite{eustratiadis2021weight}. All experiments are conducted on CIFAR-10 with the PGD attack. AT means to use adversarial training and it shows that many defense methods involve adversarial training, which requires a high computational cost. The results show that the proposed MFDV-SNN outperforms the listed defense methods greatly. It should be noted that even with some deeper network architectures, our proposed method achieves the highest accuracy on clean data.

\subsection{Inspection of Gradient Obfuscation}
\begin{figure}
  \begin{center}
  \begin{subfigure}
  {\includegraphics[scale=0.27]{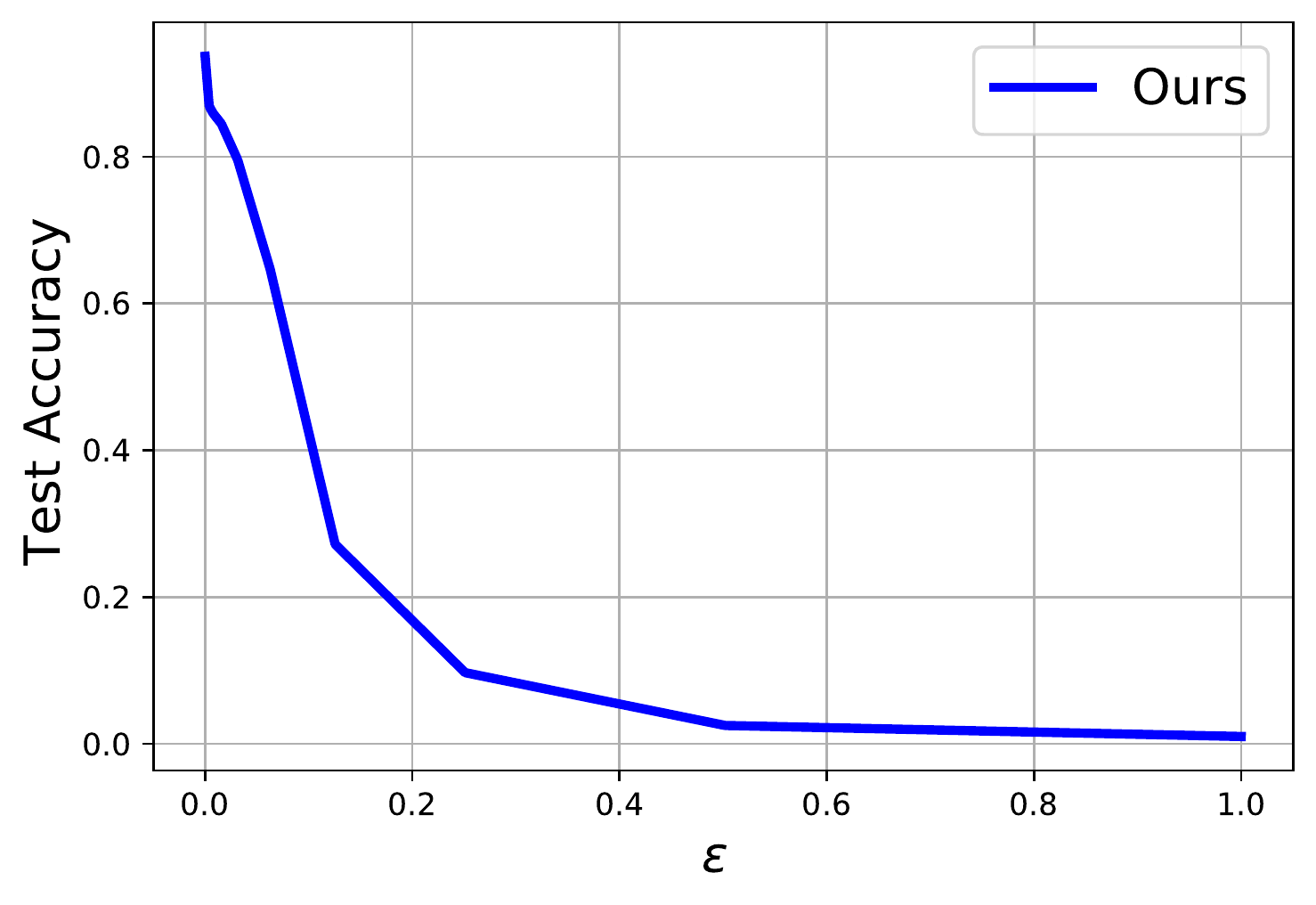}}
  \end{subfigure}
   \hfill
  \begin{subfigure}
    {\includegraphics[scale=0.27]{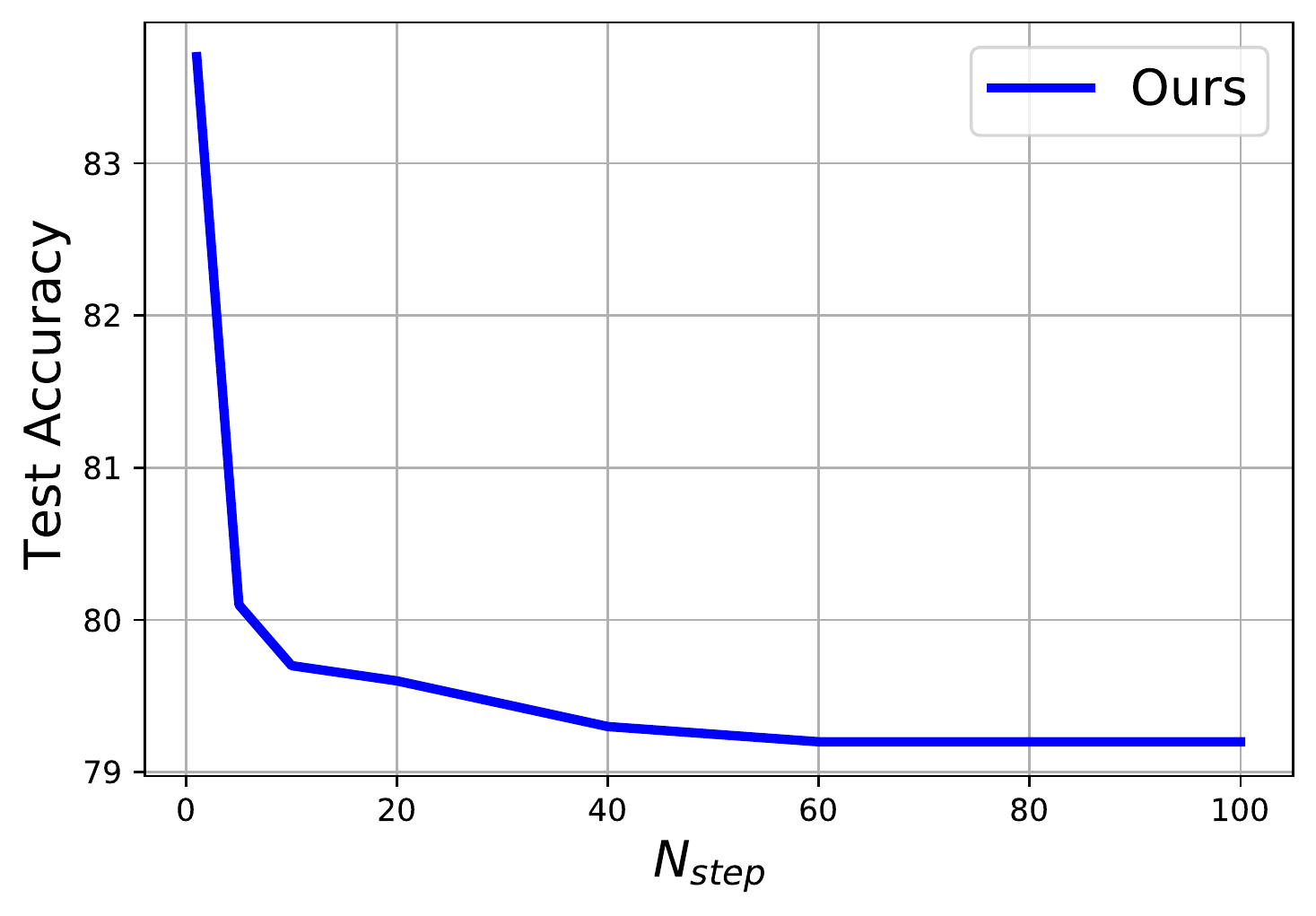}}
  \end{subfigure}
  \caption{On CIFAR-10 test set, the pertubed-data accuracy of ResNet-18 under PGD attack (Left) versus attack bound $\epsilon$, and (Right) versus number of attack steps N$_{step}$.}
  \label{gradient}
  \end{center}
\end{figure}

Athalye~\cite{athalye2018obfuscated} claimed that some stochastic algorithms are of false defense method. These methods mainly obfuscate the gradient information to improve the model's robustness. However, methods that follow obfuscated gradient can be attacked finally. So, in this section, we conduct a series of experiments to check the list proposed by Athalye~\cite{athalye2018obfuscated}, the practice experiments 
follow~\cite{jeddi2020learn2perturb}.\\
\textbf{Criterion 1:} One-step attack performs better than iterative attacks.
\\
\textbf{Refutation:} As we know, a PGD attack is an iterative attack, and FGSM is a one-step attack. It can be seen from Table~\ref{attack1} and Table~\ref{attack2} that the accuracy of MFDV-SNN against FGSM attack is consistently higher than that of PGD attack.
\\
\textbf{Criterion 2:} Black-box attacks perform better than white-box attacks.
\\
\textbf{Refutation:} From Table~\ref{attack1} and Table~\ref{pixel_attack}, the PGD attack is stronger than 1- and 2-pixel attacks, we can see that the black-box attack is worse than white-box attack.
\\
\textbf{Criterion 3:} Unbounded attacks do not reach 100$\%$ success.
\\
\textbf{Refutation:} Following~\cite{he2019parametric}, as drawn in Figure~\ref{gradient} (left), we run experiment by increasing the distortion bound-$\epsilon$. The results show that the unbounded attacks lead to 0$\%$ accuracy under attack.\\
\textbf{Criterion 4:} Random sampling finds adversarial examples.\\
\textbf{Refutation:} Following~\cite{he2019parametric}, the prerequisite is that the gradient-based (e.g., PGD and FGSM) attack cannot find adversarial examples. However, Figure~\ref{gradient} (right) indicates that when we increase the distortion bound, our method can still be broken.\\
\textbf{Criterion 5:} Increasing the distortion bound does not increase the success rate.\\
\textbf{Refutation:} The experiment in Figure~\ref{gradient} (left) shows that increasing the distortion bound increases the attack success rate. 
\\
\textbf{EOT-attack:} Athalye~\cite{athalye2018obfuscated} further claimed that the false gradient could not protect the model well when the attackers use the gradient, which is the expectation over a series of transformations. Following \cite{jeddi2020learn2perturb,pinot2019theoretical}, we use a Monto Carlo method which expects the gradient over 80 simulations of different transformations on the CIFAR-10 dataset with the backbone ResNet-18. Results show that PNI, Adv-BNN, L2P can provide 48.7$\%$, 51.2$\%$, and 53.3$\%$ robustness, respectively. The proposed MFDV-SNN can have 79.2$\%$ robustness, which outperforms the compared methods and confirms that the proposed MFDV is not of gradient obfuscation. To ensure the efficiency of the proposed MFDV-SNN result is not a stochastic gradient, we adopt 15 MC sampling at the test phase in our experiments, which number we find in experiment is stable enough.

\subsection{Inspection of Generalization}
In this section, many experiments are conducted to evaluate the generalization of the proposed model. More specifically, we first explore the dataset size influence on the proposed MFDV-SNN, as shown in Table~\ref{ablation_dataset}. Three different sizes are used in this experiment. \textbf{SVHN}: a relatively small dataset. \textbf{CIFAR-10}: a medium dataset with 60k training data and 10k testing data. \textbf{CIFAR-100}: a large dataset with one-hundred classes. The experiments are based on the backbone ResNet-18. The results show that MFDV-SNN has a great generalization to different dataset sizes.

For Table~\ref{ablation_cifar10} and Table~\ref{ablation_cifar100}, we mainly explore the impact of the network architecture on the proposed MFDV-SNN. More specifically, in practice, we explore how the depth and breadth of the network affect our method. \textbf{ResNet-18}: a common used backbone in stochastic defense. \textbf{ResNet-50}: a deeper ResNet architecture than ResNet-18. \textbf{WRN-34-10}: a wider network architecture than ResNet-18 and ResNet-50. To show the efficiency of the proposed MFDV-SNN more abundantly, we used both CIFAR-10 and CIFAR-100 datasets in our experiments. The results show that our proposed MFDV-SNN always maintains superior performance. 

\begin{table}[t]
\centering
\scalebox{0.62}
{
\begin{tabular}{ccccccccccc}
\toprule
&PGD($\epsilon/255$)&Clean&1&2&4&8&16&32&64&128\\
\midrule
\multirow{2}{*}{ResNet-18}&No Defense&92.9&62.3&32.7&10&2.3&0.2&0&0&0\\
&\textbf{MFDV-SNN}&\textbf{93.7}&\textbf{86.9}&\textbf{85.9}&\textbf{84.5}&\textbf{79.6}&\textbf{64.7}&\textbf{27.2}&\textbf{9.7}&\textbf{2.5}\\
\midrule
\multirow{2}{*}{ResNet-50}&No Defense&93.6&59.3&26.4&7.2&2.9&0.6&0&0&0\\
&\textbf{MFDV-SNN}&92.4&\textbf{83.8}&\textbf{81.7}&\textbf{79.4}&\textbf{74.1}&\textbf{56.3}&\textbf{20.9}&\textbf{7.5}&\textbf{3.6}\\
\midrule
\multirow{2}{*}{WRN-34-10}&No Defense
&94.2&54.4&23.2&11.8&6.5&1.4&0&0&0\\
&\textbf{MFDV-SNN}&\textbf93.5&\textbf{88.3}&\textbf{87.3}&\textbf{86.8}&\textbf{84.4}&\textbf{75.0}&\textbf{38.9}&\textbf{10.1}&\textbf{1.5}\\
\bottomrule
\end{tabular}}
\caption{Generalization study for PGD attacks on various architectures: ResNet-18, ResNet-50 and WRN-34-10. In which we use CIFAR-10 dataset as a backbone.}
\label{ablation_cifar10}
\end{table}

\begin{table}[t]
\centering
\scalebox{0.62}
{\begin{tabular}{ccccccccccc}
\toprule
&PGD($\epsilon/255$)&Clean&1&2&4&8&16&32&64&128\\
\midrule
\multirow{2}{*}{ResNet-18}&No Defense&68.8&32.4&15.5&5.2&1.5&0.2&0&0&0\\
&\textbf{MFDV-SNN}&\textbf{69.5}&\textbf{51.6}&\textbf{47.0}&\textbf{43.0}&\textbf{37.3}&\textbf{24.1}&\textbf{8.9}&\textbf{1.8}&\textbf{0.4}\\
\midrule
\multirow{2}{*}{ResNet-50}&No Defense&71.4&26.2&12.1&4.6&1.7&0.4&0&0&0\\
&\textbf{MFDV-SNN}&70.6&\textbf{46.2}&\textbf{40.6}&\textbf{35.5}&\textbf{25.6}&\textbf{12.9}&\textbf{3.0}&\textbf{0.5}&0\\
\midrule
\multirow{2}{*}{WRN-34-10}&No Defense
&72.0&24.7&8.5&2.0&0.6&0.1&0&0&0\\
&\textbf{MFDV-SNN}&71.2&\textbf{54.3}&\textbf{51.7}&\textbf{48.5}&\textbf{42.7}&\textbf{28.1}&\textbf{9.8}&\textbf{1.4}&\textbf{0.2}\\
\bottomrule
\end{tabular}}
\caption{Generalization study for PGD attacks on various architectures: ResNet-18, ResNet-50 and WRN-34-10. In which we use CIFAR-100 dataset as a backbone.}
\label{ablation_cifar100}
\end{table}

\begin{figure}
   \begin{center}
  \begin{subfigure}
  {\includegraphics[scale=0.27]{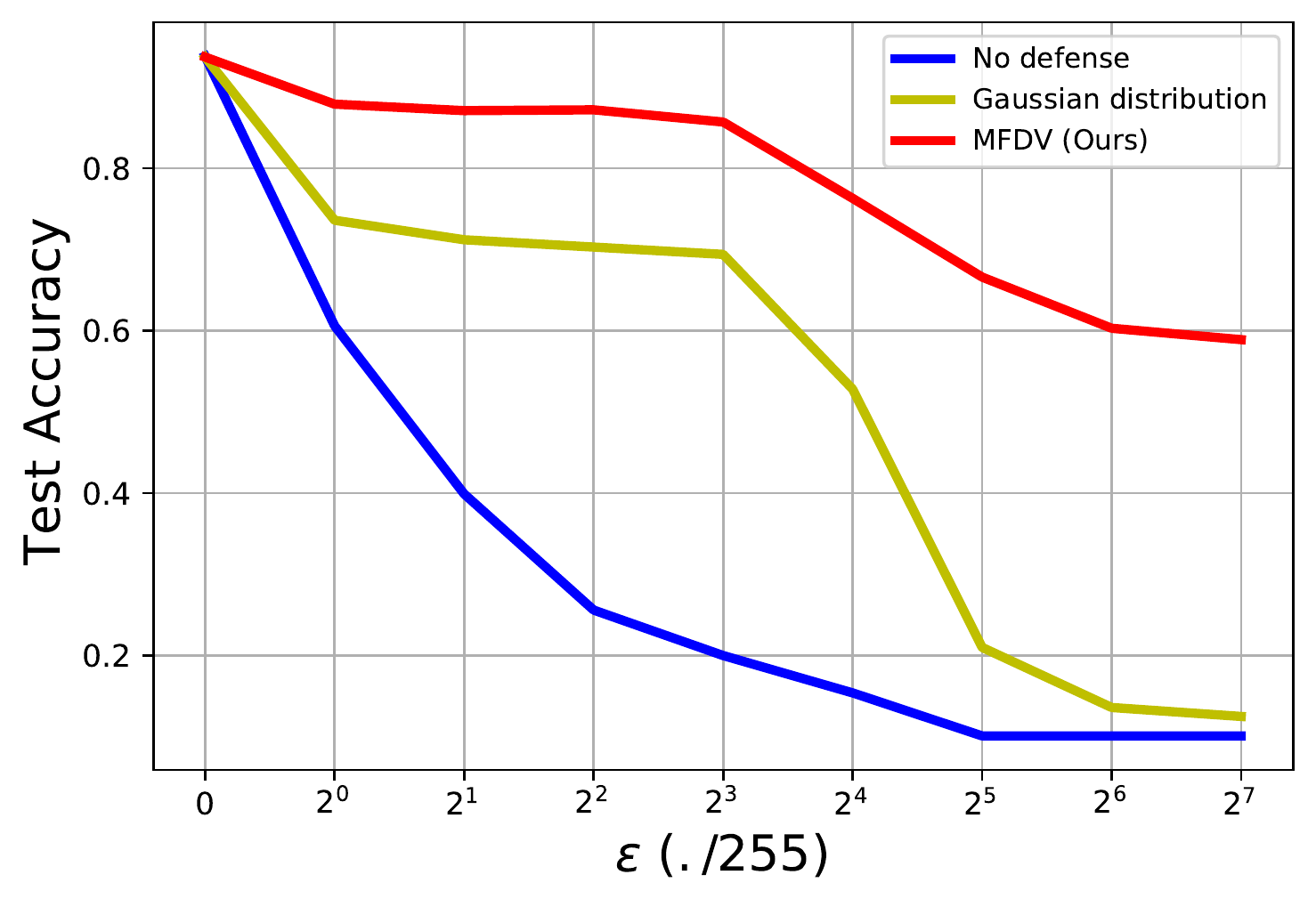}}
    \label{parametera}
  \end{subfigure}
   \hfill
  \begin{subfigure}
    {\includegraphics[scale=0.27]{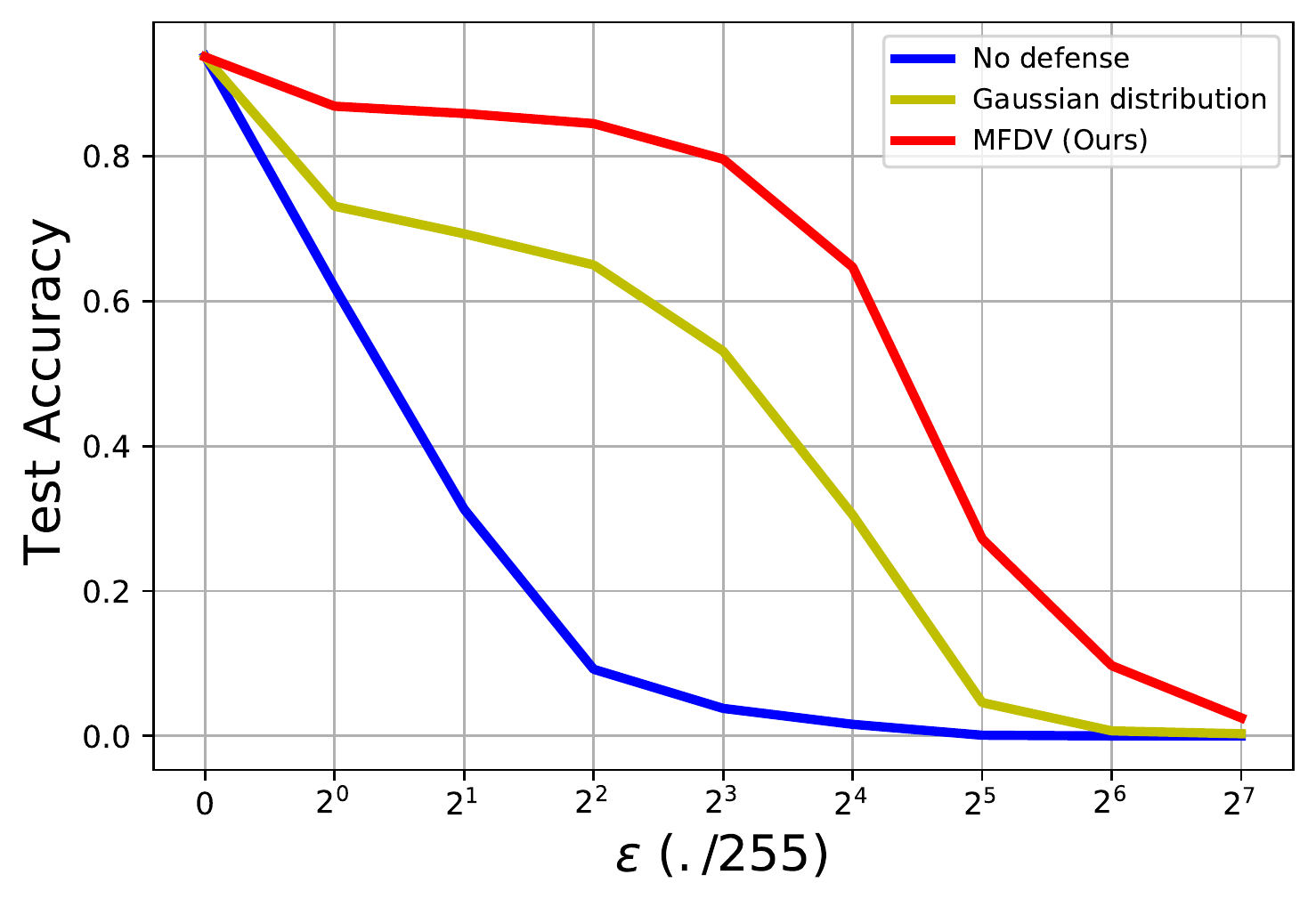}}
    \label{parameterb}
  \end{subfigure}
  \caption{Ablation study on CIFAR-10 dataset with the backbone ResNet-18. FGSM attack (Left). PGD attack (Right).}
  \label{hyperparameter}
  \end{center}
\end{figure}

\subsection{Ablation Study of MFDV-SNN}
As shown in Figure~\ref{hyperparameter}, we conduct experiments to check the efficiency of the proposed module on the FGSM (left) attack and PGD (right) attack. Note that the hyper-parameter $\lambda_{1}$ controls the power of the proposed MFDV-SNN. No defense means that we do not add randomness to the model. We train with ResNet-18 model on the CIFAR-10 dataset. Gaussian distribution means implementing the feature layer to a Gaussian distribution without regularization. Gaussian distribution with MFDV indicates that we add the critical point max feature distribution variance loss (MFDV-SNN) to the cross-entropy loss. Ablation study of MFDV-SNN shows that the proposed regularization module is effective to improve model adversarial robustness. 

\subsection{TSNE Visualization}
\begin{figure}
   \begin{center}
  \begin{subfigure}
  {\includegraphics[width=0.49\columnwidth]{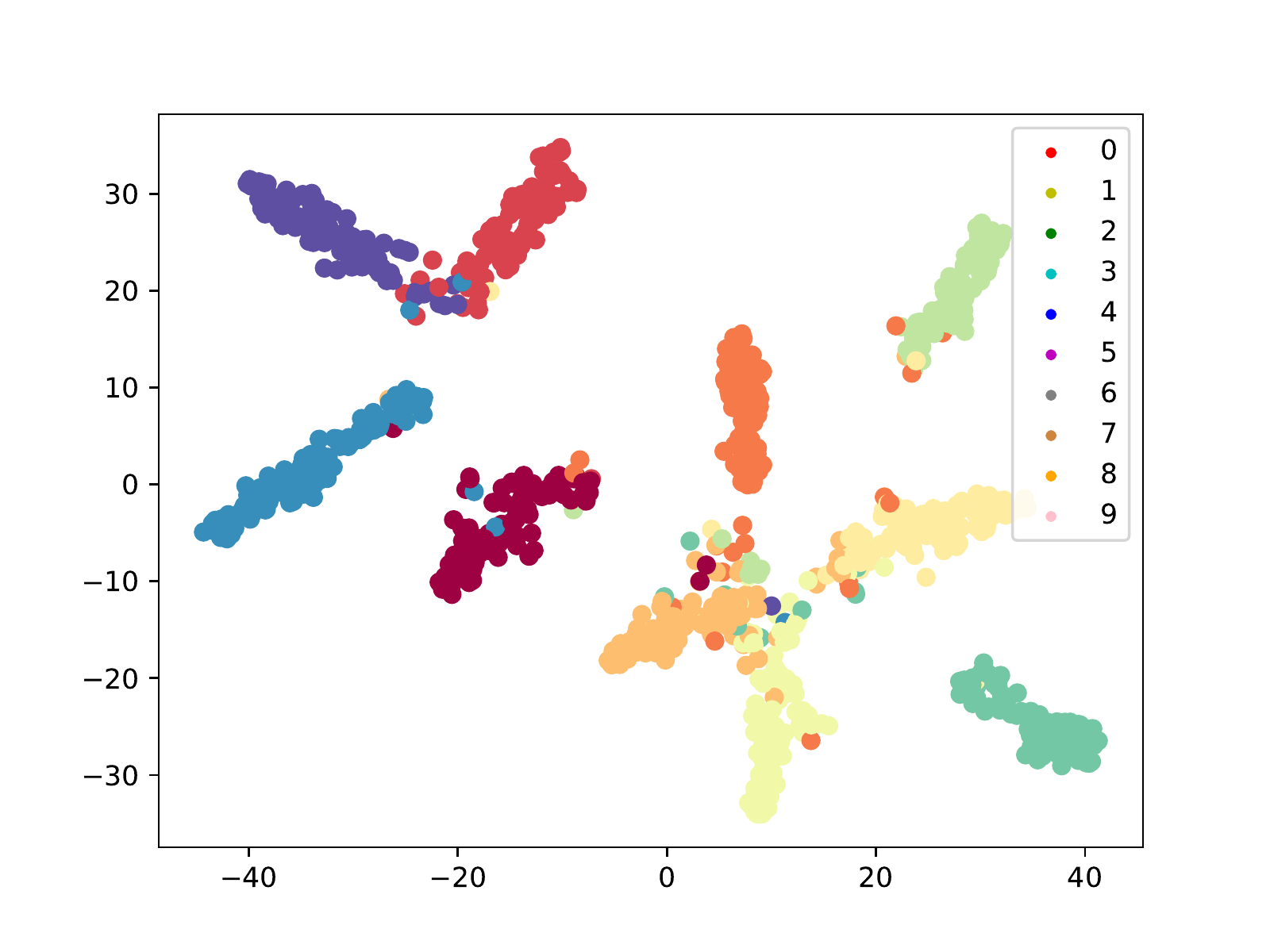}}
  \end{subfigure}
   \hfill
  \begin{subfigure}
    {\includegraphics[width=0.49\columnwidth]{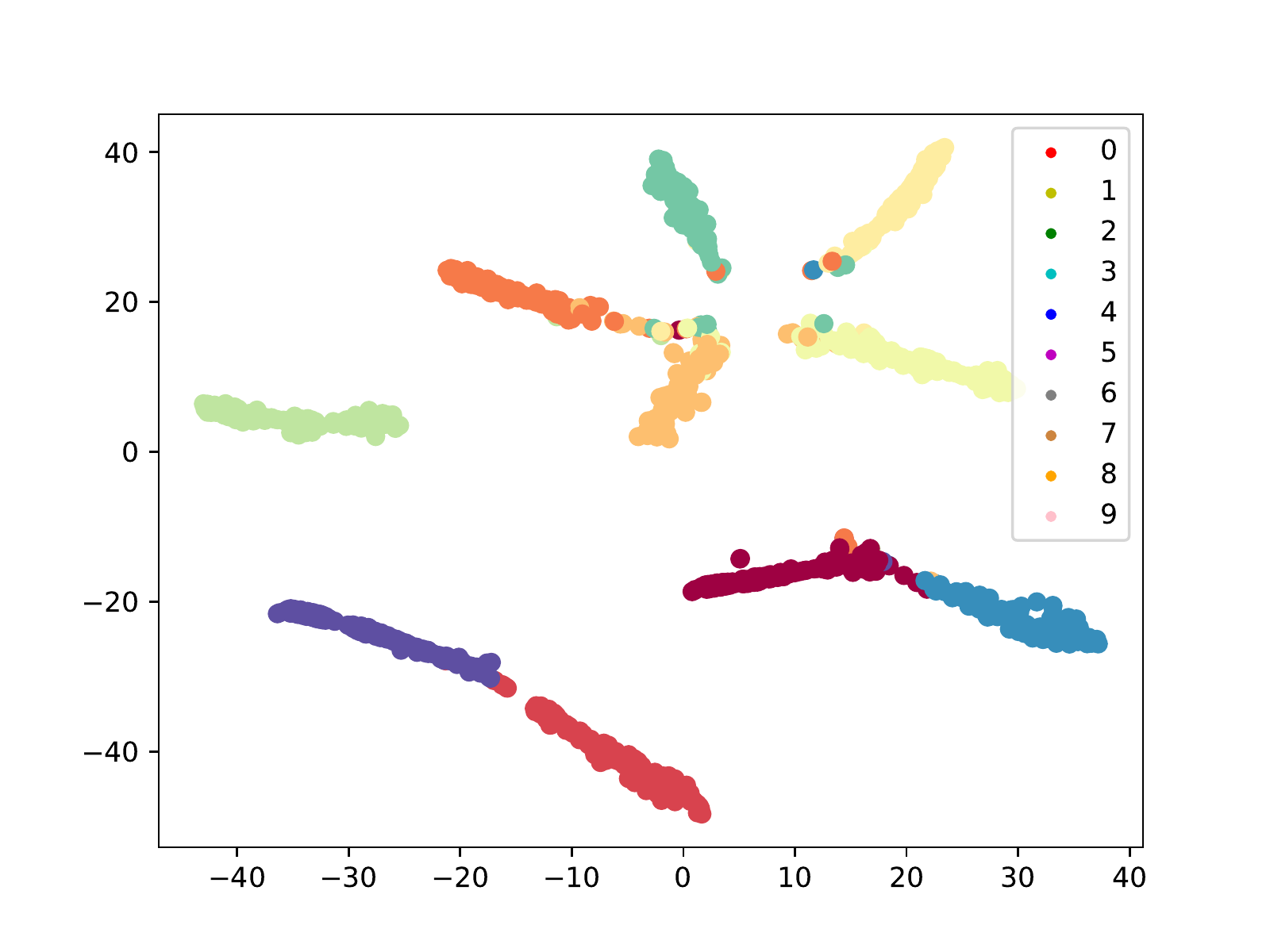}}
  \end{subfigure}
  \caption{Tsne visualization of classification result on CIFAR-10 trained on ResNet-18. No defense (Left). MFDV-SNN (Right).}
  \label{tsne}
  \end{center}
\end{figure}
We visualize the classification result on the CIFAR-10 dataset trained on ResNet-18, as shown in Figure~\ref{tsne}. In practice, we sample 1000 data and visualize them. The visualization results show that the proposed MFDV-SNN learns a more robust architecture that achieves intra-class compactness and performs better even in inter-class separation. It is proved that the proposed method, with unbounded high variance, can maintain high uncertainty and adaptively learn a deeper network representation. The uncertainty will also help the network avoid local optima and explore the global optima, thereby improving the robustness of the model and even the classification ability.

\section{Conclusion}
In this paper, we propose a simple stochastic neural network named Maximizes Feature Distribution Variance (MFDV-SNN), which significantly exceeds the existing state-of-the-art defense algorithms. Specifically, we build the feature layer to a non-informative unbounded Gaussian distribution that maximizes the Gaussian variance during model training. The proposed method does not rely on adversarial training. It has lower computation costs and training time than adversarial training-based algorithms. Extensive experiments on white- and black-box attacks show that MFDV-SNN achieves considerable performance gains and outperforms existing methods, which motivates researchers to rethink feature uncertainty for adversarial robustness in future research.
\newpage 
\appendix
\bibliographystyle{named}
\bibliography{ijcai22}

\end{document}